\documentclass[10pt,letterpaper,compsoc,conference]{iiswc26}

\usepackage{cite}
\usepackage{amsmath,amssymb,amsfonts}
\usepackage{algorithmic}
\usepackage{graphicx}
\usepackage[dvipsnames]{xcolor}
\usepackage[final]{microtype}
\usepackage[italic]{mathastext}
\usepackage{libertine}
\usepackage[T1]{fontenc}
\usepackage{textcomp}
\usepackage[varqu,varl]{zi4}
\usepackage[all]{nowidow}
\usepackage[keeplastbox]{flushend}
\usepackage{booktabs}
\usepackage{array}
\usepackage{xcolor}
\usepackage{xurl}
\usepackage{multirow}
\usepackage{tabularx}
\usepackage{placeins}
\usepackage[hidelinks]{hyperref}
\usepackage{fancyhdr}

\usepackage[table]{xcolor}
\usepackage{amssymb}
\usepackage{graphicx}

\newcommand{\tool}{\textsc{Kernel Forge }}
\newcommand{\torch}{PyTorch}
\newcommand{\eager}{\torch{} eager}

\makeatletter
\newcommand{\linebreakand}{%
  \end{@IEEEauthorhalign}%
  \hfill\mbox{}\par
  \mbox{}\hfill\begin{@IEEEauthorhalign}%
}
\makeatother

\begin{document}
\title{Kernel Forge: An Agent Harness for LLM-based\\
Generation and Optimization of CUDA Kernels}
\author{
\IEEEauthorblockN{Joshua Brodsky\IEEEauthorrefmark{1}}
\IEEEauthorblockA{University of Michigan\\
USA\\
joshbrod@umich.edu}
\and
\IEEEauthorblockN{Dhravid Kumar\IEEEauthorrefmark{1}}
\IEEEauthorblockA{University of Michigan\\
USA\\
dhravid@umich.edu}
\linebreakand
\IEEEauthorblockN{Savini Kashmira}
\IEEEauthorblockA{University of Michigan\\
USA\\
savinik@umich.edu}
\and
\IEEEauthorblockN{Jayanaka Danatanarayana}
\IEEEauthorblockA{University of Michigan\\
USA\\
jayanaka@umich.edu}
\linebreakand
\IEEEauthorblockN{Jason Mars}
\IEEEauthorblockA{University of Michigan\\
USA\\
profmars@umich.edu}
\and
\IEEEauthorblockN{Krisztian Flautner}
\IEEEauthorblockA{University of Michigan\\
USA\\
manowar@umich.edu}
\and
\IEEEauthorblockN{Lingjia Tang}
\IEEEauthorblockA{University of Michigan\\
USA\\
lingjia@umich.edu}
}
\maketitle
\begingroup
\renewcommand{\thefootnote}{\fnsymbol{footnote}}
\footnotetext[1]{Equal contribution.}
\endgroup
\pagestyle{plain}

\begin{abstract}
Machine learning models are increasingly embedded in everyday software, and
most of their runtime is spent in a small set of compute kernels such as matrix
multiplication, convolution, and normalization. Optimizing these kernels is one
of the most direct ways to reduce latency and cost, but it has traditionally
required expert engineers to hand-write low-level GPU code. Agentic systems built
on large language models (LLMs) can now generate and optimize kernels with far
less human effort, yet existing tools are largely evaluated on randomly generated
tensors and isolated kernels, emit standalone CUDA code that developers must
manually reintegrate, mostly target only LLM PyTorch models, and offer limited
support for inspecting and debugging results. We present \textsc{Kernel Forge}, an
\emph{open-source}, end-to-end agentic harness that accepts \emph{any}
unmodified PyTorch model in place. \textsc{Kernel Forge} supports vision,
diffusion, and LLM workloads, uses Monte Carlo Tree Search (MCTS) to explore multiple
optimization paths rather than a single linear refinement chain, and ships with a
graphical user interface for monitoring progress, inspecting candidate kernels, and
debugging failures. We evaluate \textsc{Kernel Forge} on four PyTorch models
spanning vision, diffusion, and LLM workloads on an NVIDIA DGX Spark with GB10 GPU. With only 50
optimization iterations per kernel, it optimizes 14 kernels to outperform PyTorch
eager mode, reaching $1.52\times$ on \texttt{adaptive\_avgpool2d} in ResNet-50,
$1.70\times$ on \texttt{group\_norm} in Stable Diffusion 3.5 Medium, $2.83\times$
on \texttt{softmax} in Gemma 4 E2B, and $1.54\times$ on \texttt{softmax} in
Qwen 3.5 35B-A3B. The code is available at: \href{https://github.com/TheJoshBrod/KernelForge}{Kernel Forge}
\end{abstract}

\section{Introduction}
\label{sec:intro}

Machine learning (ML) models are increasingly used in everyday software systems~\cite{sze2017efficient}. As a result, improving their runtime performance is important for reducing latency and computational cost. Most of a model's runtime is spent in a small set of compute kernels, such as matrix multiplication, convolution, and normalization~\cite{jouppi2017tpu,llminfgpu2025}. Optimizing these kernels is therefore one of the most direct ways to make ML systems faster and more efficient. Traditionally, this optimization has relied on expert engineers writing and tuning low-level GPU code by hand~\cite{tillet2019triton}. Although this approach can produce highly efficient implementations~\cite{dao2022flashattention}, it is slow, expensive, and requires substantial expertise.

Agentic AI systems offer a new way to approach kernel optimization. Built on large language models (LLMs), these systems can generate, revise, and evaluate candidate kernels with less direct human involvement than traditional manual tuning. Recent systems~\cite{hong2025autocomp, wang2025geak} show that this approach can produce substantial performance improvements for individual kernels. However, a kernel that performs well in isolation may not provide the same benefit when inserted into a deployed ML model. Many benchmarks evaluate correctness and latency using randomly generated tensors and isolated kernel invocations, rather than full model executions~\cite{ouyang2025kernelbench, wang2026kernelbenchx, li2025tritonbench, wen2025multikernelbench, zhu2026cudabench, backendbench2025}. By contrast, in real workloads, kernels run under model-specific conditions, including tensor shapes, activation distributions, neighboring operators, and memory behavior that may be absent from isolated-kernel benchmarks. Hence, there is a need for tooling that bridges the gap between optimizing kernels in isolation and evaluating their behavior within real model executions on real workloads.

Promising tools, including some open-source systems, have begun to address this problem~\cite{flashinferbench2026, cudaforge2025,astra2025,lange2025robustkbench,hong2025autocomp, kernelagent2025, ouyang2025kernelbench}. However, they still leave several challenges unresolved for applying kernel optimization to real PyTorch models. 

First, existing tools do not make model-level optimization easy for developers without low-level CUDA expertise. Applying kernel optimization to real PyTorch workloads requires more than generating a fast standalone kernel: the optimized code must be inserted into the model and validated for correctness. Many current tools instead leave this integration step to the developer~\cite{cudaforge2025,astra2025,lange2025robustkbench}. Second, some tools focus primarily on LLM workloads or benchmark kernels, while real PyTorch applications include diffusion, vision, and audio models that use the same operators with different shapes, layouts, and arguments~\cite{flashinferbench2026,ouyang2025kernelbench,astra2025}. Because kernel performance depends on these model-specific details, optimizations tuned for one workload may not transfer to others. Third, existing tools may not explore alternative optimization paths effectively. Linear refinement can make results depend heavily on early design choices~\cite{cudaforge2025,astra2025}, while beam-style search can prune paths that appear less promising early, leading to suboptimal solutions in heuristic search~\cite{zhang1998completebeam}. This is challenging because useful optimization paths may pass through temporarily slower intermediate candidates. Finally, many systems are exposed mainly as research prototypes, scripts, or
benchmark workflows rather than integrated graphical tools. This makes kernel
inspection, progress tracking, candidate comparison, failure diagnosis, and
application to different PyTorch models more manual~\cite{cudaforge2025,
astra2025,lange2025robustkbench,kernelagent2025}. Together, these limitations
motivate an open-source, end-to-end system for applying, inspecting, and
evaluating kernel optimizations within real PyTorch workloads.

\begin{figure*}[t]
    \centering
    \includegraphics[width=\textwidth]{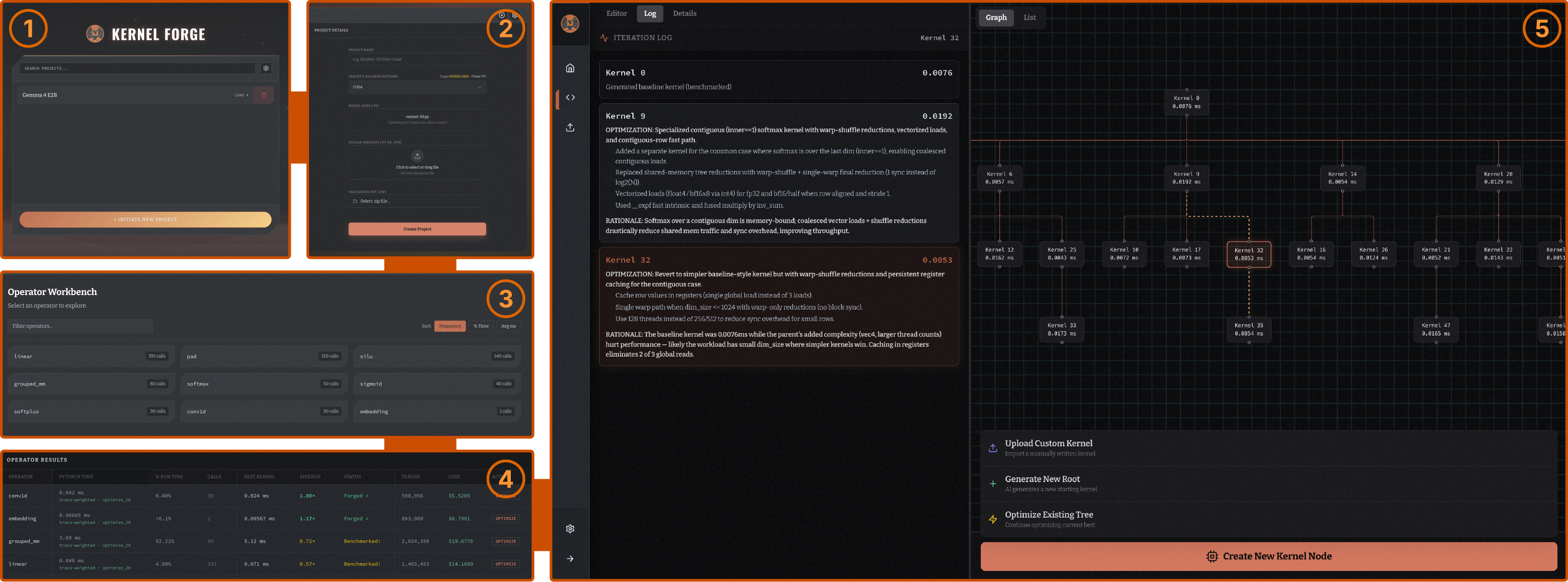}
    \caption{\tool graphical interface. 
    The numbered screenshots show the main workflow:
    (1) the project launcher lists saved optimization projects;
    (2) project creation lets users load models and weights directly from Hugging Face, configure the target backend, and provide validation inputs without modifying model code;
    (3) the operator workbench shows captured kernels, their runtime prominence, and their optimization status so users can choose what to optimize;
    (4) the results view summarizes per-operator timing, speedup, cost, and backend or fallback status;
    and (5) the kernel workbench exposes generated CUDA revisions, agent reasoning, optimization decisions, and the MCTS revision tree for each optimization run.}
    \label{fig:gui-overview}
\end{figure*}

To address these limitations, we introduce \textsc{Kernel Forge}, an open-source agentic AI harness for optimizing real PyTorch workloads beyond LLMs. \textsc{Kernel Forge} accepts any standard PyTorch model, along with user-provided example inputs, without requiring users to modify the model code. Unlike tools that output optimized kernels as standalone code, \textsc{Kernel Forge} validates generated kernels against PyTorch eager outputs and automatically integrates them back into the PyTorch model execution path. It supports diverse PyTorch workloads, including language, diffusion, vision, and audio models, rather than being limited to a single workload family. To explore optimization paths more flexibly than linear refinement or beam-style search, \textsc{Kernel Forge} uses Monte Carlo Tree Search (MCTS)~\cite{kocsis2006uct}, allowing it to revisit candidates that may not show immediate gains. As shown in Figure~\ref{fig:gui-overview}, to improve observability and usability, \textsc{Kernel Forge} provides both a graphical user interface (GUI) and a command-line interface (CLI) for interacting with the optimization process. Users can easily monitor optimization progress, inspect candidate kernels, debug failures, and select among different model providers to balance cost, performance, and deployment needs.


\textsc{Kernel Forge} works by running the model on the provided example inputs and capturing the PyTorch operators that execute during inference. It then attaches to high-level operators such as \texttt{conv2d}, \texttt{matmul}, \texttt{group\_norm}, and \texttt{softmax}, and uses an agentic loop to propose, generate, validate, and profile specialized hardware-aware CUDA kernels. Generated kernels are numerically validated against PyTorch eager outputs on the captured workload  and execute transparently within eager-mode model execution.

We evaluate \textsc{Kernel Forge} on four PyTorch models representing vision, diffusion, and LLM workloads: ResNet-50~\cite{he2016resnet}, Stable Diffusion 3.5 Medium~\cite{stability2024sd35}, Gemma 4 E2B~\cite{google_gemma4_model_card}, and Qwen 3.5 35B-A3B~\cite{qwen35_model_card}. All experiments run on an NVIDIA DGX Spark with GB10 GPU~\cite{nvidia_dgx_spark_hw}. With only 50 optimization iterations per kernel, \textsc{Kernel Forge} produces kernels that outperform PyTorch eager mode across all four models. It reaches $1.52\times$ eager-mode performance for \texttt{adaptive\_avgpool2d} in ResNet-50, $1.70\times$ for \texttt{group\_norm} in Stable Diffusion 3.5 Medium, $2.83\times$ for \texttt{softmax} in Gemma 4 E2B, and $1.54\times$ for \texttt{softmax} in Qwen 3.5 35B-A3B. These results show that LLM-driven kernel optimization can be applied not only to isolated benchmarks, but also to operators captured from full PyTorch models and representative workloads.

In this paper, we make the following contributions:
\begin{enumerate}
    \item We present \textsc{Kernel Forge}, an open-source, end-to-end tool that optimizes any real PyTorch model executions by capturing executed operators, generating specialized CUDA kernels, exploring multiple optimization paths, and integrating optimized kernels back into the model execution path.

    \item We provide a GUI and CLI for monitoring optimization progress, inspecting generated kernels, debugging failures, and selecting among different model providers.

    \item We evaluate \textsc{Kernel Forge} on four PyTorch models representing vision, diffusion, and LLM workloads, showing up to $2.83\times$ eager-mode performance on selected operators with only 50 optimization iterations per kernel.

    \item We release \textsc{Kernel Forge} as an open-source tool to support reproducible research and practical adoption of agentic kernel optimization for PyTorch workloads.
\end{enumerate}

\begin{figure*}[t]
    \centering
    \includegraphics[width=\textwidth]{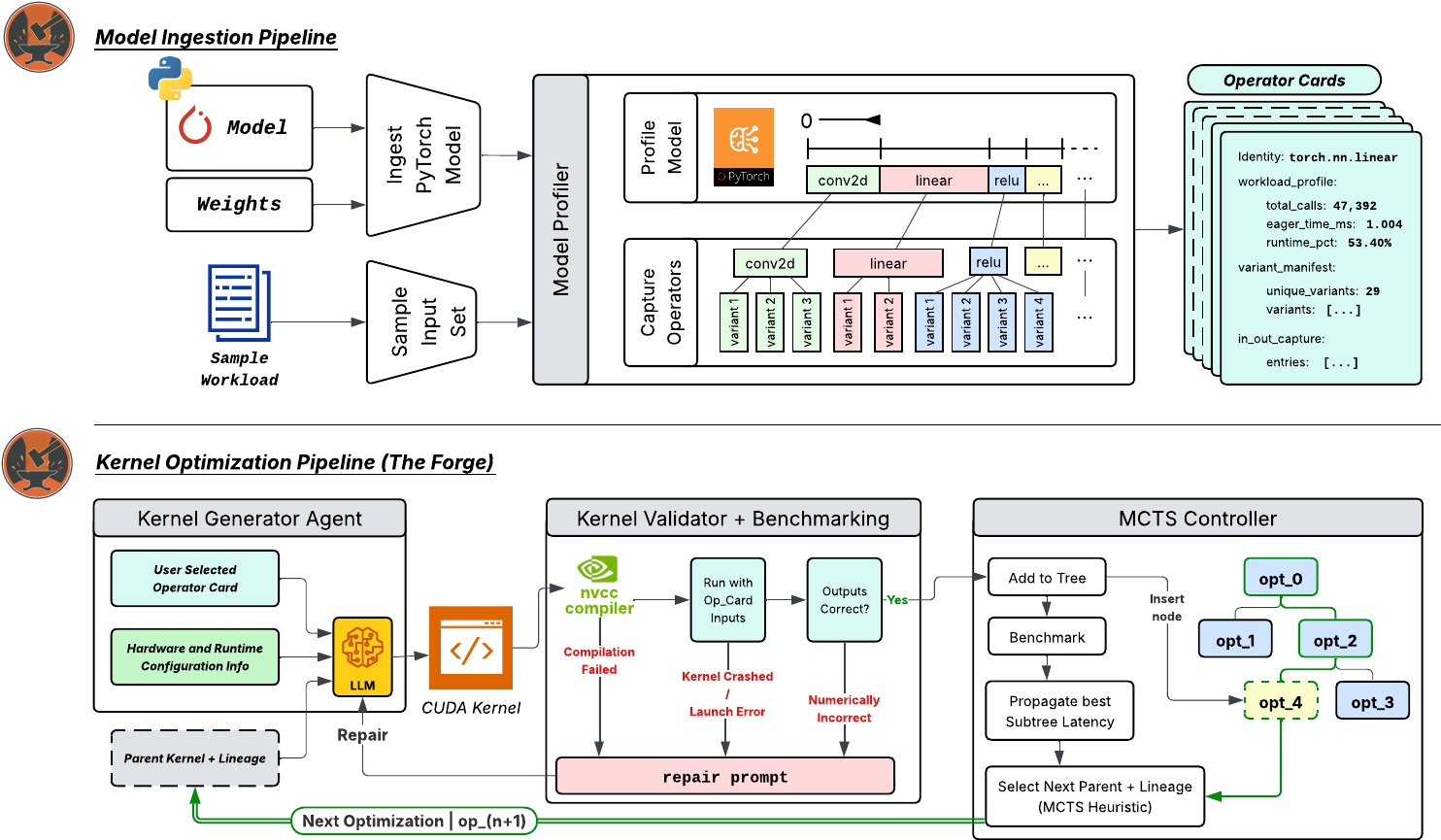}
    \caption{Kernel Forge system overview. The model ingestion pipeline captures workload-specific PyTorch operator variants as operator cards, and the Forge uses those cards to generate, validate, benchmark, and iteratively optimize CUDA candidates with an MCTS controller.}
    \label{fig:system}
\end{figure*}

\section{Background}
\label{sec:background}
\subsection{PyTorch Operator Capture for Kernel Optimization}
PyTorch models are written using high-level operators such as matrix
multiplication, convolution, normalization, and softmax~\cite{paszke2019pytorch}.
During inference, these operators call lower-level GPU kernels that perform the
actual computation. The efficiency of each kernel then depends on the tensors and
arguments passed to the operator at runtime, including shapes, layouts, and data
types.

A standalone benchmark may test a kernel using randomly generated tensors or a
small set of fixed shapes~\cite{ouyang2025kernelbench}. In contrast, a real PyTorch model may invoke the same
operator with different shapes, layouts, arguments, and surrounding memory
behavior. Therefore, it is useful to observe which operators execute in the
model and optimize kernels using the inputs observed during inference.
Once the relevant operators and inputs are known, the optimizer can generate
specialized kernels for those cases. The generated kernels can then be validated
against PyTorch eager outputs on the captured workload and used within the model
execution path.

\subsection{GPU Kernel Optimization}
GPU kernels implement the low-level computations behind PyTorch operators.
For NVIDIA GPUs, these kernels are commonly written in CUDA, which gives
developers control over thread organization, memory access, synchronization, and
data movement~\cite{nvidia_cuda_programming_guide}. Optimizing a kernel often
requires hardware-aware choices such as tiling, memory coalescing,
shared-memory reuse, operator fusion, and reducing synchronization or launch
overhead~\cite{nvidia_cuda_best_practices}.

These choices can substantially affect performance, but they are difficult to
make by hand. A developer must preserve the behavior of the original operator
while reasoning about GPU hardware details and workload-specific inputs. Existing
libraries, compilers, and domain-specific languages (DSL) such as TVM and Triton help generate efficient
implementations~\cite{chen2018tvm,chen2018autotvm,zheng2020ansor,tillet2019triton}, but specialized kernels can
still be useful when a model invokes an operator with particular shapes,
arguments, or memory behavior~\cite{dao2022flashattention}.

\subsection{Agentic Kernel Generation}
Recent systems use LLMs to generate and improve GPU kernels
through an iterative feedback loop~\cite{hong2025autocomp,wang2025geak}.
In a typical loop, the system generates a candidate kernel, compiles and runs it,
checks its output against a reference implementation, measures performance, and
uses the result to guide later attempts. This makes it possible to improve
kernels over multiple iterations rather than relying on a single generated
implementation.

A key challenge is deciding which candidates to keep exploring. Existing tools
often use linear refinement, which improves one candidate through a sequence of
edits~\cite{cudaforge2025,astra2025}, or beam-style search, which keeps several candidates but may prioritize
immediate performance gains~\cite{hong2025autocomp}. In kernel optimization, this can be limiting: a
candidate that is slower in an early iteration may become useful after later
changes, while a candidate that is fast on one test case may not generalize to
the model workload. Monte Carlo Tree Search (MCTS) is useful in this setting
because it can explore multiple optimization paths while using correctness and
performance feedback to focus search on promising directions~\cite{kocsis2006uct}.
Practical kernel optimization therefore requires code generation, correctness
checking, profiling, and tools for inspecting the optimization process.

\section{Kernel Forge System Design}
\label{sec:system}

\textsc{Kernel Forge} is an end-to-end harness for optimizing PyTorch operators
that execute during real model inference. Given a PyTorch model, its weights,
and example inputs, \textsc{Kernel Forge} captures supported operator calls,
groups them into workload-specific variants, and optimizes selected variants
with generated CUDA kernels. Each candidate is validated and benchmarked against
PyTorch eager execution, then refined through a persistent MCTS search. The final
output is a guarded package that dispatches generated CUDA only when the captured
measurements show it is valid and faster; otherwise, execution falls back to the
original PyTorch eager path.

Figure~\ref{fig:system} shows the two main stages of the system. \S~\ref{sec:model-ingestion} describes how the model ingestion pipeline turns a real PyTorch workload into operator cards. Sections~\ref{sec:kernel-generator}--\ref{sec:mcts-controller} describe how the Forge turns a selected operator card into generated CUDA candidates and iteratively improves them through validation, benchmarking, repair, and search. \S~\ref{sec:packaging} describes how optimized candidates are exported with guarded dispatch and audit labels.

\subsection{Model Ingestion Pipeline}
\label{sec:model-ingestion}

As shown in Figure~\ref{fig:system}, \textsc{Kernel Forge} begins with a PyTorch
model, its weights, and user-provided example inputs. The user does not need to
rewrite the model or manually identify which kernels to optimize. Instead,
\textsc{Kernel Forge} executes the original model on the provided inputs and
records the PyTorch operators that run during inference.

The model profiler records supported operators such as \texttt{conv2d},
\texttt{linear}, \texttt{relu}, normalization, pooling, and \texttt{softmax}.
For each operator call, it captures the information needed to replay and
validate that call, including tensor shapes, dtypes, strides, layouts, scalar and
optional arguments, device placement, output metadata, and example inputs and
outputs. This allows later stages to optimize the operator behavior that actually
appears in the workload, rather than a synthetic standalone kernel.

\textsc{Kernel Forge} then groups captured calls into runtime-compatible
variants, as illustrated by the variant boxes in Figure~\ref{fig:system}. A
variant is a concrete operator case, not just a public operator name. Two calls
to the same operator may form different variants if they differ in shape, dtype,
layout, arguments, backend behavior, or replacement requirements. Variants become
the units passed to the optimization pipeline.

The output of ingestion is a set of operator cards, shown on the right side of
the ingestion pipeline in Figure~\ref{fig:system}. Each card summarizes a
captured operator family, its call count, eager execution time, runtime share,
variant manifest, and validation examples. These cards define the concrete
optimization targets while preserving the PyTorch eager execution as the
reference for later correctness and performance comparisons.

\subsection{Kernel Generator Agent}
\label{sec:kernel-generator}




As shown in the Forge portion of Figure~\ref{fig:system}, the kernel generator
agent consumes an operator card from the model ingestion pipeline and produces a
CUDA candidate for the selected variant. The operator card defines the concrete
target for generation: tensor metadata, argument values, output metadata, and the
launch interface used for validation and benchmarking. The generator also
receives the hardware and runtime configuration so that candidates can be
specialized to the target GPU and execution environment.

For the first candidate, the generator prompts an LLM to produce CUDA source and
a launch wrapper for the selected variant. This candidate becomes the first node
in the optimization sequence. In later rounds, the search controller selects a
previously validated parent kernel from the MCTS tree and passes its source and
lineage back to the generator. The generator then proposes a revised candidate
intended to reduce measured latency while preserving the behavior captured by the
operator card.

The generation prompt asks the LLM to produce custom CUDA code rather than
delegating the computation to PyTorch, ATen, cuDNN, cuBLAS, or other framework
and library implementations. \textsc{Kernel Forge} does not assume that the model
always follows this instruction: later validation and auditing stages label each
candidate so that generated CUDA, backend-wrapper, and fallback paths can be
distinguished in the final results.

\subsection{Kernel Validator}
\label{sec:kernel-validation}

As shown in the Forge portion of Figure~\ref{fig:system}, each CUDA candidate
produced by the generator must pass validation before it can enter the MCTS
tree. \textsc{Kernel Forge} first compiles the generated source with
\texttt{nvcc}. If compilation fails, the compiler error is returned to the
generator through a bounded repair loop, and the repaired candidate is compiled
again.

After compilation, \textsc{Kernel Forge} launches the candidate using the inputs
stored in the operator card. Runtime failures, such as launch errors, crashes, or
illegal memory accesses, are also sent to the repair loop with the corresponding
failure information. Candidates that compile and launch successfully are then
checked against the PyTorch eager outputs captured during model ingestion, using
the numerical tolerance configured for the operator and dtype.

Validation is a workload-specific filter rather than a proof of correctness for
all possible inputs. A passing candidate matches the captured PyTorch outputs for
the observed workload, but may not be correct outside those cases. The validator
therefore decides only whether a candidate is admissible to the search tree.
Performance ranking and export decisions are handled later by the MCTS
controller and guarded packaging policy.

\begin{table*}[t]
  \centering
  \caption{Workloads, capture sources, and timing policy used in the evaluation. Captured calls are operator-region calls used for trace weighting.}
  \label{tab:methods}
  \scriptsize
  \setlength{\tabcolsep}{3.0pt}
  \begin{tabularx}{\textwidth}{>{\raggedright\arraybackslash}p{0.13\textwidth}>{\raggedright\arraybackslash}p{0.23\textwidth}>{\raggedright\arraybackslash}p{0.27\textwidth}>{\raggedright\arraybackslash}X r}
  \toprule
  Workload & Model / checkpoint & Execution and capture source & Timing policy & Captured calls \\
  \midrule

  ResNet-50 &
  TorchVision ResNet-50 V1.5 with \texttt{IMAGENET1K\_V2} weights~\cite{he2016resnet,torchvision_resnet50_weights}, 16-bit floating point (fp16) CUDA. &
  ImageNetV2 matched-frequency images~\cite{recht2019imagenetv2} for workload context; operator-region evaluation includes captured convolution, normalization, residual-add, activation, pooling, flatten, and linear operators. &
  224$\times$224 preprocessing; 20 warmup batches and 100 timed batches for reported batch contexts. &
  2{,}800 \\

  \addlinespace[0.35em]

  SD3.5 Medium &
  Stability AI Stable Diffusion 3.5 Medium~\cite{stability2024sd35} in Diffusers~\cite{von-platen-etal-2022-diffusers}, bfloat16 (bf16) CUDA. &
  T2I-CompBench text-to-image prompts~\cite{huang2023t2icompbench} for workload context; operator-region evaluation includes captured linear, attention, normalization, activation, convolution, embedding, elementwise-add, softmax, padding, and interpolation operators. &
  Batch size 1, 1024$\times$1024, 28 denoising steps, guidance 3.5, max sequence length 256; 20 warmup and 100 timed prompt generations. &
  75{,}544 \\

  \addlinespace[0.35em]

  Gemma 4 E2B &
  \texttt{google/gemma-4-E2B-it}~\cite{google_gemma4_model_card}, bf16, CUDA/offload, eager attention. &
  Fixed 100-prompt language-model set: 60 prompts from MT-Bench~\cite{zheng2023mtbench}, 20 prompts from the ShareGPT Prompts Annotated dataset~\cite{lewtun2023sharegptprompts}, and 20 prompts from the LongBench \texttt{multi\_news} subset~\cite{bai2024longbench}. &
  Batch size 1, greedy decoding, key-value (KV) cache enabled, 128 new-token limit; 20 warmup and 100 timed generations. &
  3{,}840 \\

  \addlinespace[0.35em]

  Qwen 3.5 35B-A3B &
  \texttt{Qwen/Qwen3.5-35B-A3B}~\cite{qwen35_model_card}, bf16, CUDA/offload, eager attention. &
  Fixed 100-prompt language-model set: 60 prompts from MT-Bench~\cite{zheng2023mtbench}, 20 prompts from the ShareGPT Prompts Annotated dataset~\cite{lewtun2023sharegptprompts}, and 20 prompts from the LongBench \texttt{multi\_news} subset~\cite{bai2024longbench}. &
  Batch size 1, greedy decoding, key-value (KV) cache enabled, 128 new-token limit; 20 warmup and 100 timed generations. &
  912 \\

  \bottomrule
  \end{tabularx}
\end{table*}

\subsection{MCTS Controller}
\label{sec:mcts-controller}

As shown in the Forge portion of Figure~\ref{fig:system}, validated CUDA
candidates are passed to the MCTS controller for benchmarking and tree updates.
The controller maintains a persistent revision tree for each captured variant.
Each node represents a candidate kernel and stores its source, parent and child
links, visit count, optimization notes, audit status, and measured latency when
available.

After a candidate passes validation, \textsc{Kernel Forge} benchmarks it against
the corresponding PyTorch eager path for the same variant and captured inputs.
The benchmark records candidate latency, eager latency, source type, and status
labels. These measurements are local to the captured variant; full-model effects
are evaluated separately.

The controller updates each node with its measured latency and propagates the
best latency observed in the node's subtree through the ancestor chain. The tree
retains both faster and slower valid candidates because a slower candidate may
still provide a useful starting point for later revisions. The updated tree is
then used to select the next parent kernel and lineage sent back to the generator.

Selection combines progressive widening with a minimization version of Upper
Confidence bounds applied to Trees (UCT)~\cite{kocsis2006uct}. For a node $x$
with $N_x$ visits, the controller may generate another child while
\begin{equation}
\left|\mathrm{children}(x)\right| \le
\left\lfloor N_x^{\alpha(R)} \right\rfloor ,
\end{equation}
where $R$ is the root visit count. We anneal $\alpha(R)$ from $0.5$ to $0.3$ over
the first 1000 root visits, allowing broader branching early in search and more
reuse of existing branches later.

When a node is not expanded, the controller descends to the child with the
lowest score:
\begin{equation}
\mathrm{score}(c \mid p)
= L_c - C\sqrt{\frac{\log N_p}{N_c}},
\end{equation}
where $L_c$ is the child's best-subtree latency when available and otherwise its
own measured latency, $N_p$ and $N_c$ are the parent and child visit counts, and
$C=1.0$ in our configuration. This objective minimizes measured latency for the
selected captured variant while still encouraging exploration of less-visited
branches.

\subsection{Packaging, Guarded Dispatch, and Audit Labels}
\label{sec:packaging}

After optimization by the controller in \S~\ref{sec:mcts-controller},
\textsc{Kernel Forge} can export selected candidates into \texttt{.cast} packages.
Each package contains the generated source, launch metadata, validation status,
timing information, workload-specific operator metadata, and audit labels needed
for guarded runtime replacement.

The export policy is guarded: a candidate is used only when it is available,
compatible with the captured call, numerically valid on the recorded workload
examples, and faster than the measured PyTorch eager latency for the same
variant. Otherwise, the package falls back to the original PyTorch eager path.
This allows \textsc{Kernel Forge} to substitute generated CUDA only where the
captured evidence supports a benefit, while preserving strong framework or
vendor-backed implementations when PyTorch eager remains faster or more reliable.

\textsc{Kernel Forge} also records whether each selected path is custom generated
CUDA, a wrapper or backend path, or a fallback to PyTorch eager. This audit trail
separates three outcomes that are easy to conflate: whether the package improves
the captured operator-region result, whether the selected path uses generated
CUDA, and whether fallback was needed to preserve the eager baseline. Wrapper or
backend paths are included in operator-region accounting because they affect the
exported package, but they are excluded from generated-CUDA-win claims.

The resulting package does not uniformly replace PyTorch eager execution.
Instead, guarded dispatch makes a replacement decision at the captured-variant
level. If the generated CUDA candidate is valid, compatible with the observed
call pattern, and faster than the measured PyTorch eager path, the package
dispatches the generated kernel. Otherwise, it preserves the original PyTorch
implementation.

This behavior is important because real PyTorch workloads contain a mix of
operators. Some operators expose opportunities for specialized generated CUDA,
while others already dispatch to highly optimized framework or vendor-backed
implementations. Guarded dispatch allows \textsc{Kernel Forge} to apply generated
kernels only where the captured evidence supports a benefit, while avoiding
regressions where PyTorch eager remains the stronger choice.

Together, the ingestion pipeline, generator, validator, MCTS controller, and
guarded package form an end-to-end workflow for safely applying agentic kernel
optimization to real PyTorch model executions. We next evaluate how this workflow
performs across representative PyTorch workloads.

Figure~\ref{fig:gui-overview} shows the graphical interface exposed around
this workflow. The interface supports project creation, kernel inspection,
tree-based optimization monitoring, queue tracking, LLM usage accounting, and
per-operator result inspection. These views are not part of the measured runtime
path; they provide the observability needed to apply and debug the optimization
workflow on real PyTorch projects.

\section{Evaluation}
\label{sec:results}

\subsection{Evaluation Methodology}
\label{sec:evaluation}


We evaluate \tool{} to understand how agentic kernel optimization behaves when applied to operators captured from real PyTorch model executions. Unlike isolated kernel benchmarks, this setting compares generated kernels against the actual PyTorch eager paths used by the workload, including mature framework and vendor-backed implementations. Our evaluation focuses on three questions: (i) whether \tool{} can generate valid kernels that outperform the PyTorch eager path for selected captured operators, (ii) where these wins occur relative to each operator's runtime contribution, and (iii) how search cost changes across optimization budgets.

The evaluation uses four PyTorch workloads covering vision, diffusion, and LLM
inference: ResNet-50 for vision, Stable Diffusion 3.5 Medium for diffusion, and
Gemma 4 E2B and Qwen 3.5 35B-A3B for LLM inference. All experiments run on an
NVIDIA DGX Spark. The evaluation stack uses PyTorch 2.10.0+cu130, CUDA 13.0
with \texttt{nvcc} 13.0 V13.0.88, NVIDIA driver 580.126.09, cuDNN 9.15.1, and
an NVIDIA GB10 GPU reporting compute capability 12.1 (\texttt{sm\_121}).
Table~\ref{tab:methods} summarizes the model checkpoints, workload inputs,
timing policy, and captured operator-region size for each workload. All CUDA
generation and optimization prompts issued by the agent use Anthropic Claude
Opus 4.7; validation, benchmarking, guarded fallback, and cost accounting are
performed by \tool{}'s local harness.

For each workload, \tool{} first runs the unmodified PyTorch model on representative inputs and records the captured PyTorch operators that execute during inference. Each captured variant corresponds to a concrete operator invocation pattern observed in the workload, not merely an operator family. A variant includes the operator name, tensor shapes, dtypes, layouts, non-tensor arguments, reference outputs, call count, and measured PyTorch eager latency. We use the observed call counts and eager timings to compute each variant's runtime percentage within the captured operator region. This weighting is important because a large speedup on a rarely executed operator has less deployment impact than a smaller improvement on an operator that dominates runtime.

The reported results are operator-region measurements. By focusing on the captured operator region, we isolate the part of execution where \tool{} can make a replacement decision while still weighting results by the operator mix observed in real model executions.

\subsection{Baselines, Reporting Policy, and Cost Accounting}
\label{sec:baselines_reporting_cost}

The baseline for every captured operator variant is the PyTorch eager implementation observed in the original workload. Generated kernels are therefore compared against the actual PyTorch path used for the same input shape, dtype, layout, and argument configuration, including framework or vendor-backed implementations such as cuDNN, cuBLAS, attention kernels, and ATen. This makes the comparison deployment-relevant, but also means that generated CUDA must often compete against mature library paths.

We use the guarded export policy described in \S~\ref{sec:packaging} when interpreting replacement outcomes and fallback behavior. In the figures, bars below 1.0$\times$ show generated candidates that are slower than eager, while translucent caps show the performance recovered by fallback. The caps are not additional generated-kernel speedup; they visualize the effect of the guard. We also use the audit labels from \S~\ref{sec:packaging} to distinguish custom generated CUDA from wrapper or backend paths when reporting generated-CUDA wins.

 We report five search arms. We use \emph{opt0} to denote the initial generated candidate and \emph{opt\_i} to denote the candidates which have undergone \emph{i} optimization iterations. The \emph{opt5}, \emph{opt10}, \emph{opt20}, and \emph{opt50} arms are budgeted audit points after 5, 10, 20, and 50 optimization iterations for a captured variant. These arms are not a guarantee of monotonic improvement in aggregate operator-region performance. Additional search can discover better candidates, but it can also spend budget on variants with little runtime percentage or on operators whose PyTorch eager baselines are already highly optimized.

Cost accounting records the Opus 4.7 API spend associated with each arm.
We convert input and output token usage to dollars using Anthropic's published
Claude Opus 4.7 API pricing at the time of measurement: \$5 per million input
tokens and \$25 per million output tokens~\cite{anthropic_opus47_pricing}.
We report both incremental cost, which measures the additional spend incurred
within a given arm, and cumulative cost, which measures total spend accumulated
through that arm. Because the operator-region outcome is runtime-percentage-weighted,
cost is most meaningful when considered together with runtime percentage and
baseline strength. \S~\ref{sec:search_cost} reports this accounting across
workloads and search budgets.

\subsection{Per-Model Opt50 Operator Results}
\label{sec:per_model_opt50_results}

\begin{figure}[!t]
\centering
\includegraphics[width=\columnwidth]{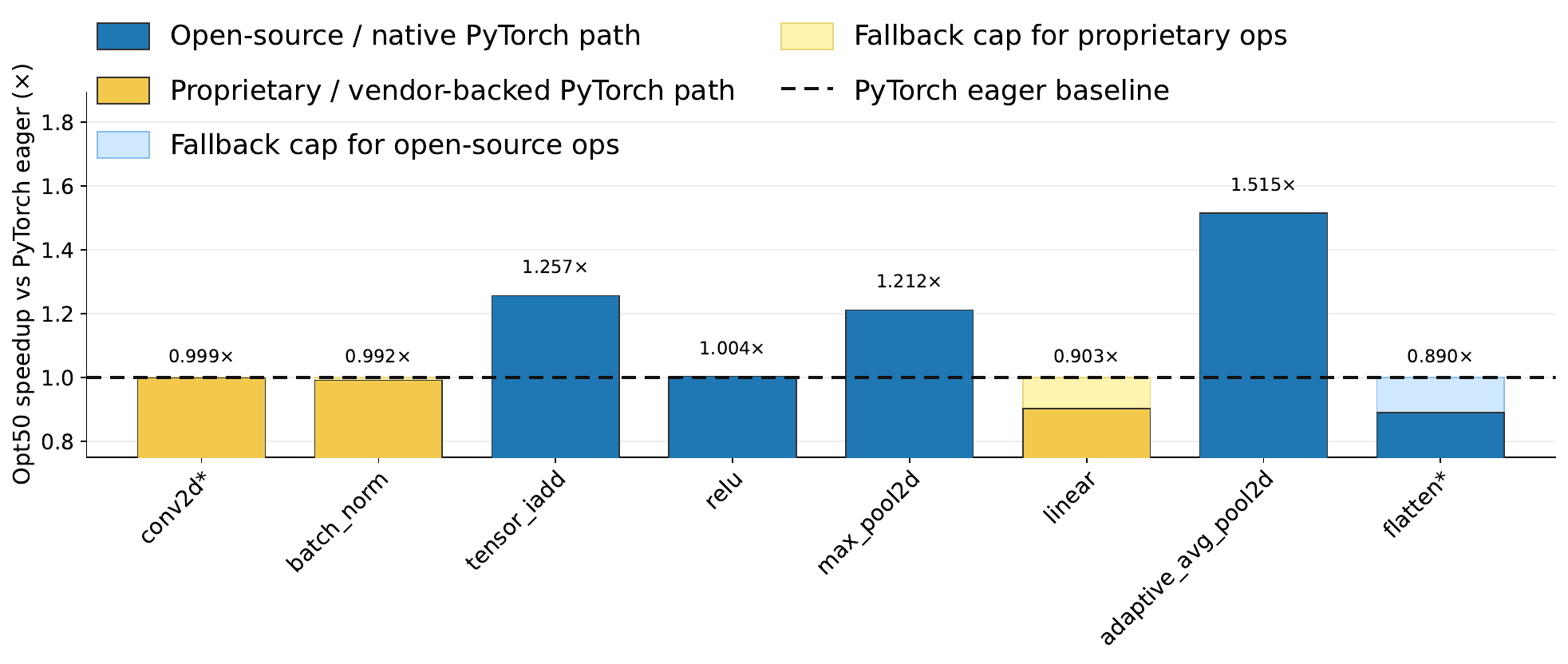}
\caption{ResNet-50 opt50 operator-level results. Bars report generated-candidate speedup relative to \eager{}; translucent caps show guarded fallback to 1.0$\times$. Asterisks mark wrapper/backend operators excluded from generated-CUDA-win claims.}
\label{fig:resnet50_opt50_results}
\end{figure}

\begin{figure}[!t]
\centering
\includegraphics[width=\columnwidth]{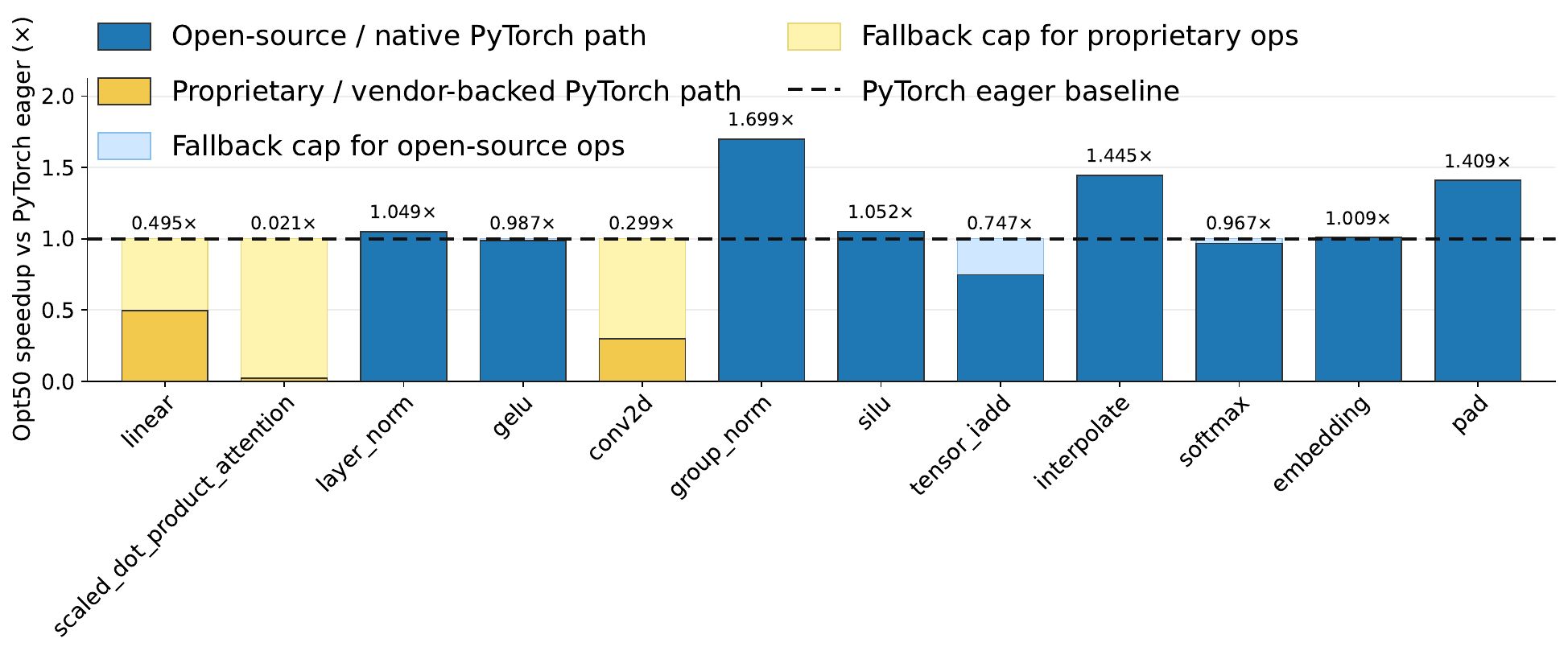}
\caption{Stable Diffusion 3.5 Medium opt50 operator-level results. Bars report generated-candidate speedup relative to \eager{}; translucent caps show guarded fallback to 1.0$\times$.}
\label{fig:sd35_opt50_results}
\end{figure}

\begin{figure}[!t]
\centering
\includegraphics[width=\columnwidth]{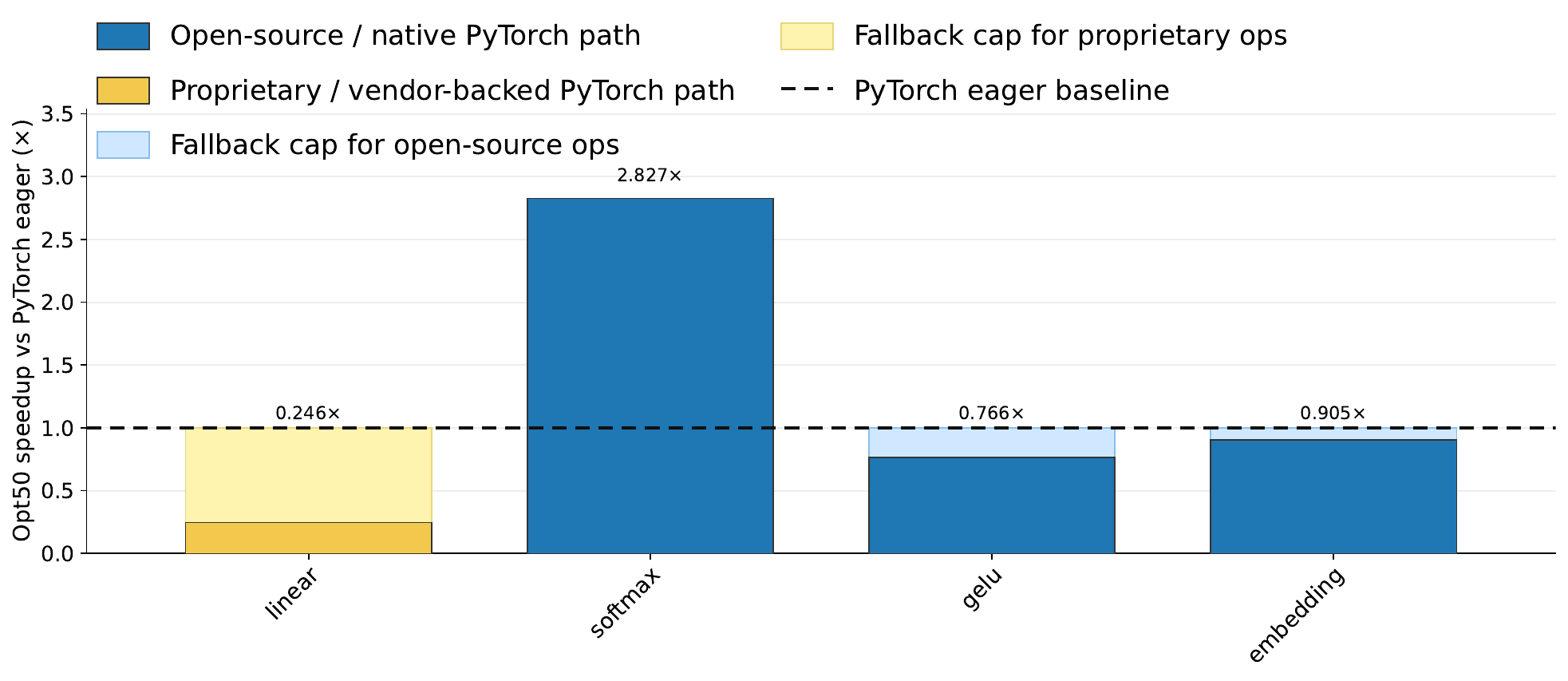}
\caption{Gemma 4 E2B opt50 operator-level results. Bars report generated-candidate speedup relative to \eager{}; translucent caps show guarded fallback to 1.0$\times$.}
\label{fig:gemma4_opt50_results}
\end{figure}

\begin{figure}[!t]
\centering
\includegraphics[width=\columnwidth]{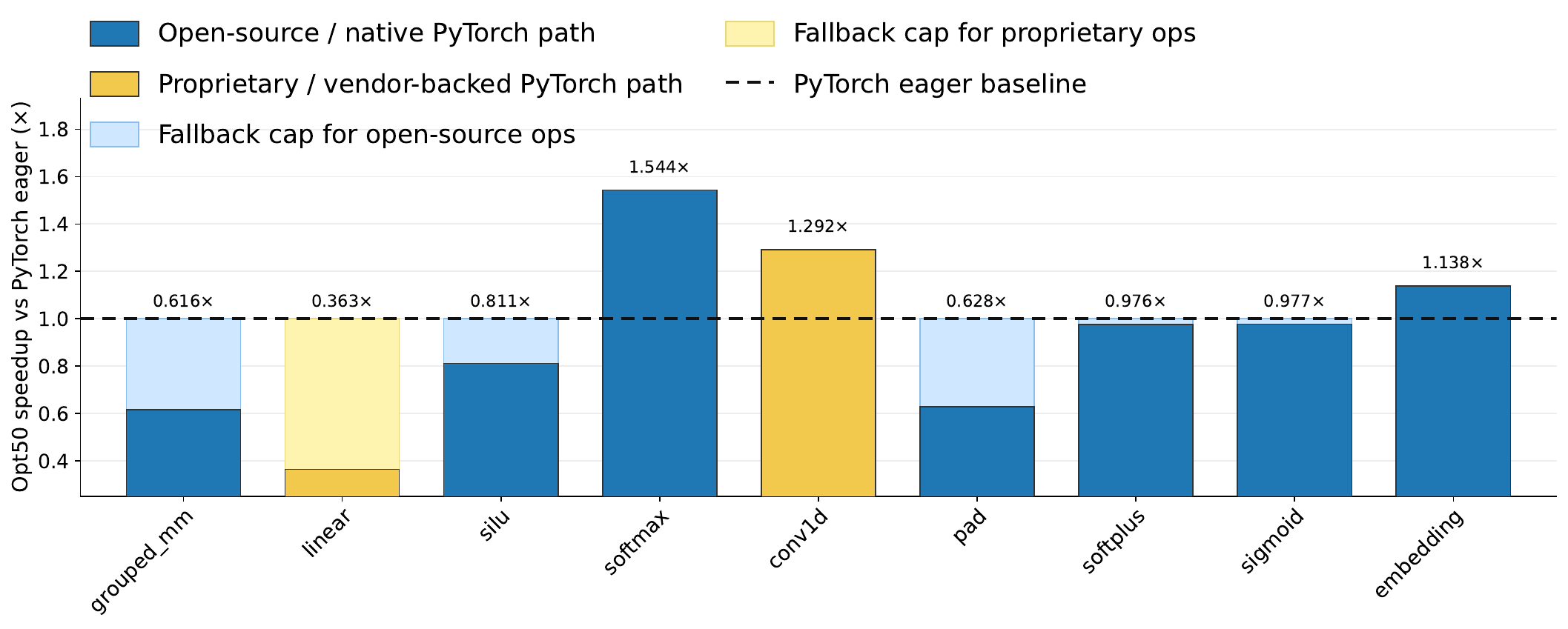}
\caption{Qwen 3.5 35B-A3B opt50 operator-level results. Bars report generated-candidate speedup relative to \eager{}; translucent caps show guarded fallback to 1.0$\times$.}
\label{fig:qwen35_opt50_results}
\end{figure}

Figures~\ref{fig:resnet50_opt50_results}--\ref{fig:qwen35_opt50_results} report the opt50 operator-level results for each workload. Across these operators, generated candidates exceed \eager{} more frequently on open-source/native PyTorch paths than on proprietary/vendor-backed PyTorch paths. Among the displayed opt50 operators, 13 of 24 open-source/native operators are faster than \eager{}, compared with 1 of 9 proprietary/vendor-backed operators. This distinction is important because the highest-runtime-percentage operators are often backed by mature framework or vendor implementations. Generated CUDA candidates are therefore compared against the actual PyTorch eager paths used by the workload, not against weak reference kernels.

Each bar gives the generated candidate speedup relative to the corresponding \eager{} operator measurement. The dashed line marks the 1.0$\times$ \eager{} baseline. Bars below 1.0$\times$ are generated candidates that are slower than \eager{}; translucent caps indicate the performance recovered by guarded fallback and are not additional generated-kernel speedup.

In ResNet-50, the generated candidates that exceed \eager{} are concentrated in open-source/native operators. Tensor add, ReLU, max pooling, and adaptive average pooling reach 1.257$\times$, 1.004$\times$, 1.212$\times$, and 1.515$\times$, respectively, with runtime percentages of 13.71\%, 11.90\%, 1.63\%, and 0.12\%. The proprietary/vendor-backed operators do not exceed \eager{}: conv2d reaches 0.999$\times$, batch normalization reaches 0.992$\times$, and linear reaches 0.903$\times$. Since conv2d alone accounts for 53.35\% of captured operator-region time and is marked as a wrapper/backend operator rather than a generated-CUDA win, fallback is necessary to preserve the dominant \eager{} path while still using generated kernels for smaller native operators.

Stable Diffusion 3.5 Medium shows the same pattern more sharply. Several open-source/native operators improve over \eager{} at opt50: group normalization reaches 1.699$\times$, layer normalization reaches 1.049$\times$, and SiLU reaches 1.052$\times$. However, these successful operators are not the dominant runtime contributors: group normalization, layer normalization, and SiLU together account for 10.48\% of captured operator-region time. The largest operators remain below \eager{}: linear reaches 0.495$\times$ at 53.40\% runtime percentage, scaled-dot-product attention reaches 0.021$\times$ at 27.12\% runtime percentage, and conv2d reaches 0.299$\times$ at 2.55\% runtime percentage. Linear and scaled-dot-product attention together account for 80.52\% of the captured region, so guarded fallback prevents slower generated candidates from replacing the most consequential \eager{} implementations.

Gemma 4 E2B is dominated by a single high-runtime-percentage operator family. Softmax reaches 2.827$\times$ at opt50, indicating that \tool{} can find large local improvements for an open-source/native operator. However, softmax accounts for only 5.93\% of captured operator-region time. Linear accounts for 90.13\% of the captured region and remains at 0.246$\times$. The guarded policy therefore preserves \eager{} for linear while retaining the generated softmax improvement.

Qwen 3.5 35B-A3B follows the same structure. Softmax and conv1d improve over \eager{}, reaching 1.544$\times$ and 1.292$\times$, while embedding reaches 1.138$\times$. These operators have small captured runtime percentages: softmax accounts for 0.19\%, conv1d accounts for 0.40\%, and embedding is approximately 0.002\% of the captured region. In contrast, grouped matrix multiplication accounts for 93.62\% of captured operator-region time and remains at 0.616$\times$. Linear also remains below \eager{} at 0.363$\times$ and accounts for 4.98\%. Guarded fallback therefore preserves the dominant PyTorch eager operators and admits generated kernels only for operators where the generated implementation is faster.

Overall, the per-model results support two conclusions. First, LLM-generated CUDA can produce substantial operator-level speedups on captured operators from real PyTorch workloads. Second, the operators most successfully optimized at opt50 often account for a small fraction of captured runtime, while the highest-runtime-percentage operators are frequently backed by mature framework or vendor implementations. This does not prove that proprietary/vendor-backed operators are inherently impossible to optimize, but it is consistent with the expectation that heavily used library paths have stronger human-engineered baselines. \tool{} addresses this deployment setting through guarded selection: it uses generated kernels where they improve measured operator latency and retains \eager{} where they do not.

\begin{figure}[!t]
\centering
\includegraphics[width=\columnwidth]{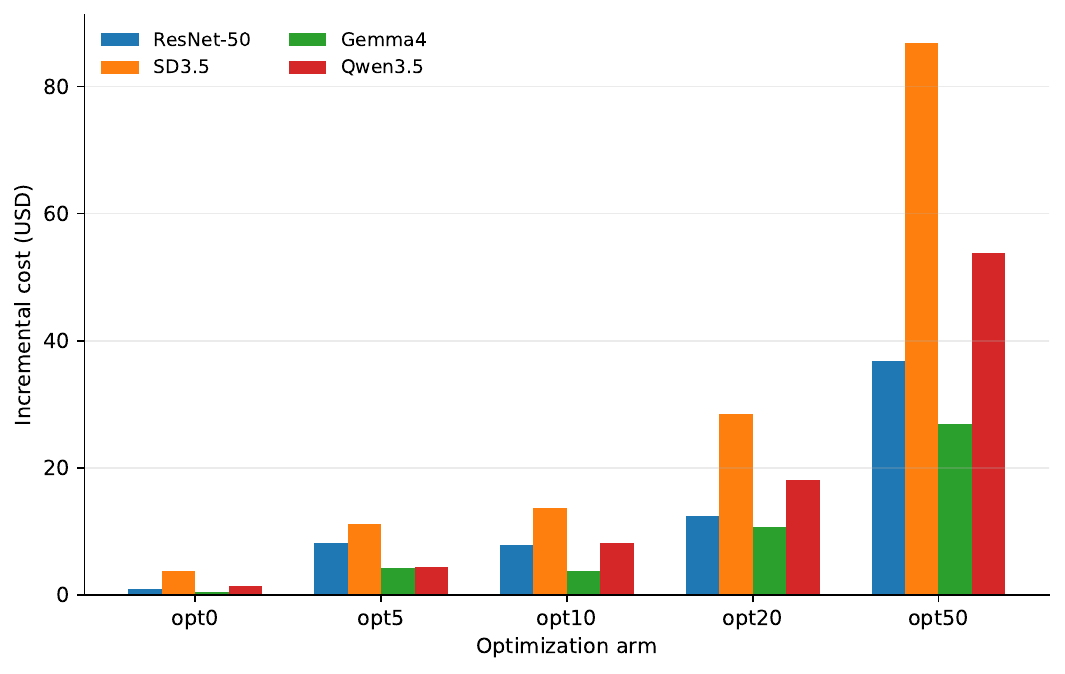}
\caption{Incremental LLM API cost by optimization arm. Bars report the additional Opus 4.7 API spend incurred within each arm for each workload; costs are not cumulative.}
\label{fig:cost_incremental_by_arm}
\end{figure}

\begin{figure}[!t]
\centering
\includegraphics[width=\columnwidth]{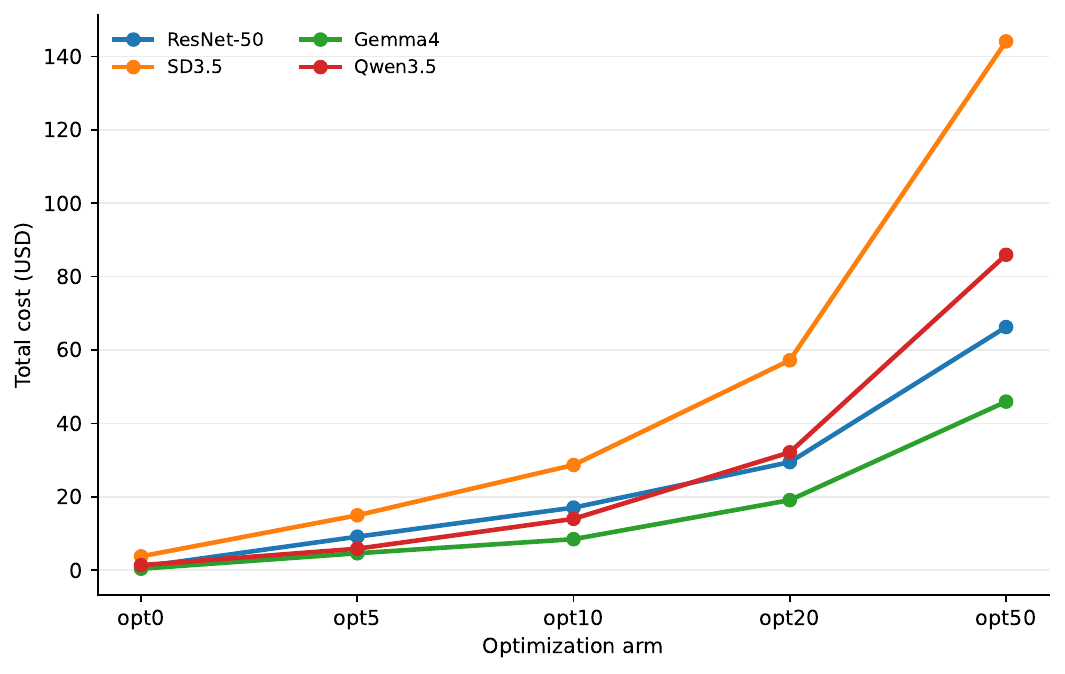}
\caption{Cumulative LLM API cost by optimization arm. Points report recorded Opus 4.7 API spend accumulated through each arm for each workload, so the opt50 point includes spend from opt0 through opt50.}
\label{fig:cost_cumulative_by_arm}
\end{figure}

\subsection{Search Cost}
\label{sec:search_cost}

Figures~\ref{fig:cost_incremental_by_arm} and~\ref{fig:cost_cumulative_by_arm} report the recorded LLM API cost across all optimized captured operators. The incremental view in Figure~\ref{fig:cost_incremental_by_arm} shows where new spend is introduced, while the cumulative view in Figure~\ref{fig:cost_cumulative_by_arm} shows the total spend accumulated by the time each search arm is reached. Across workloads, cost rises sharply at opt50 because this arm adds many more candidate-generation and repair attempts than the earlier budgets.

The cost increase is not uniformly matched by runtime-percentage-weighted benefit. In the opt50 operator results above, several of the largest local speedups occur on operators that account for a small fraction of captured operator-region time. Conversely, the dominant operators in each workload are often framework or vendor-backed paths where generated candidates remain below \eager{}. This means that spending more search budget can improve individual variants without necessarily improving the operators that determine the aggregate operator-region outcome.

\begin{figure}[!t]
\centering
\includegraphics[width=\columnwidth]{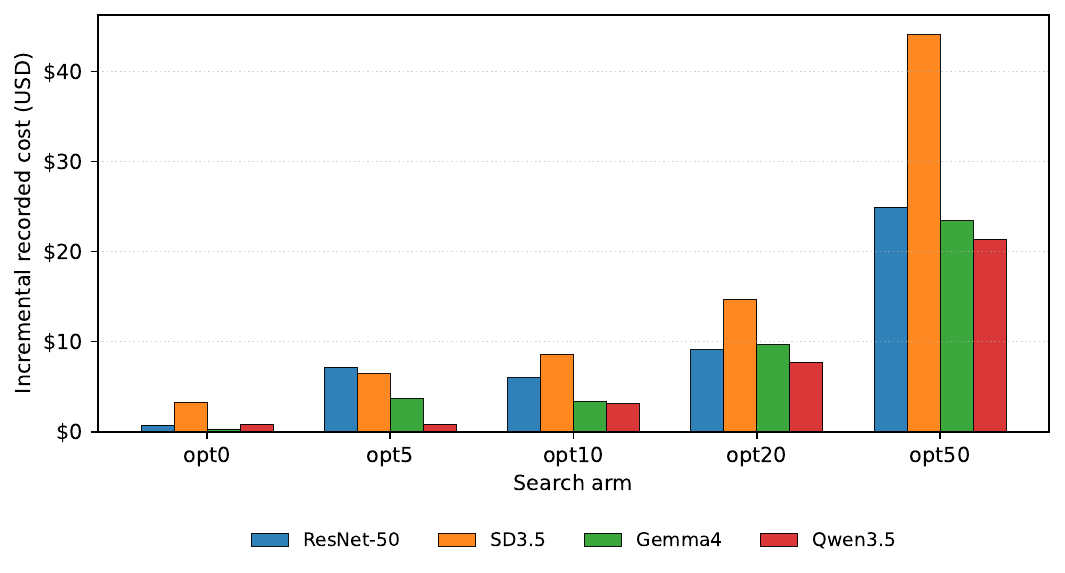}
\caption{Incremental LLM API cost by optimization arm for operators with at least 1\% captured runtime percentage. Bars report the additional Opus 4.7 API spend incurred within each arm after filtering to operators that account for at least 1\% of operator-region runtime; costs are not cumulative.}
\label{fig:cost_incremental_geq1_runtime}
\end{figure}

\begin{figure}[!t]
\centering
\includegraphics[width=\columnwidth]{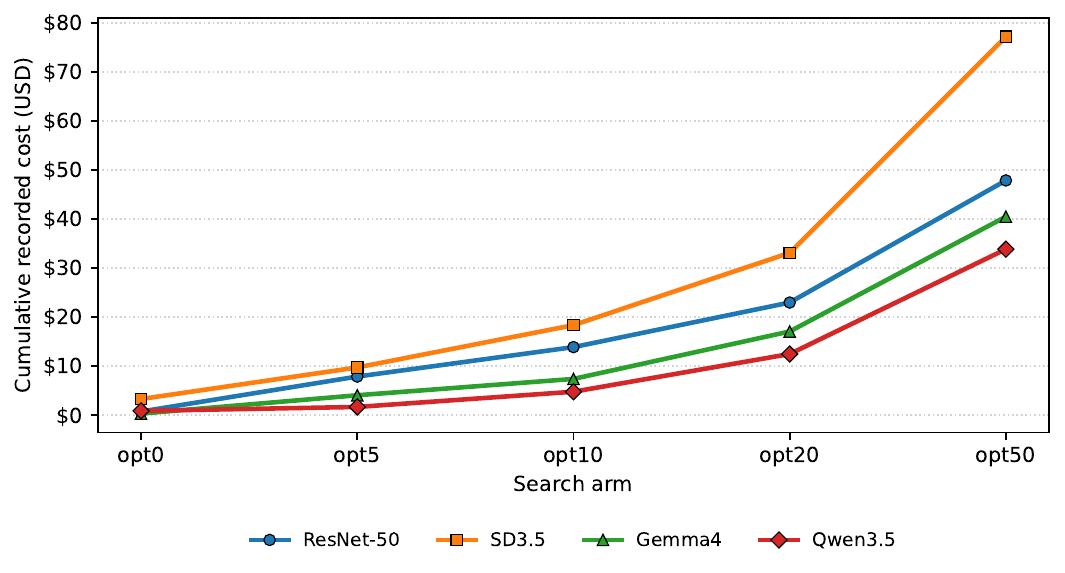}
\caption{Cumulative LLM API cost by optimization arm for operators with at least 1\% captured runtime percentage. Points report recorded Opus 4.7 API spend accumulated through each arm after filtering to operators that account for at least 1\% of operator-region runtime.}
\label{fig:cost_cumulative_geq1_runtime}
\end{figure}

Figures~\ref{fig:cost_incremental_geq1_runtime} and~\ref{fig:cost_cumulative_geq1_runtime} repeat the same accounting after filtering to operators responsible for at least 1\% of captured operator-region runtime. This filtered view removes very low-runtime-percentage operators and focuses the cost analysis on variants that could plausibly affect the aggregate result. The same pattern remains: later search arms, especially opt50, account for a large fraction of total spend. The main conclusion is not that larger search budgets are useless, but that fixed per-operator budgets are poorly aligned with deployment impact when high-runtime-percentage operators are already served by strong PyTorch backends.

These results suggest that future versions of \tool{} should allocate search budget according to runtime percentage and baseline strength rather than applying the same optimization depth to every captured variant. A runtime-percentage-aware policy could stop early on low-impact operators, spend more selectively on dominant operators, and avoid continuing search when repeated candidates fail to approach the PyTorch eager baseline.

\section{Related Work}
\label{sec:related}

\definecolor{lightgreen}{RGB}{204,255,204}
\definecolor{lightred}{RGB}{255,204,204}
\newcommand{\yes}{\cellcolor{lightgreen}$\checkmark$}
\newcommand{\no}{\cellcolor{lightred}}

\definecolor{lightyellow}{RGB}{255,245,180}
\newcommand{\partialmark}{\cellcolor{lightyellow}$\triangle$}

\begin{table*}[t]
\centering
\caption{Comparison of \tool{} with related LLM-based kernel optimization systems on the challenge. 
A check mark indicates that the system directly addresses the challenge, a triangle indicates partial support, and a blank red cell indicates that the challenge is not a primary goal of the system.}
\label{tab:tool-comparison}
\resizebox{\textwidth}{!}{
\begin{tabular}{|p{7.5cm}|c|c|c|c|c|c|c|c|}
\hline
\textbf{Capability (addresses challenge from Section~\ref{sec:intro})} 
& \textbf{Astra} 
& \textbf{GEAK} 
& \textbf{CudaForge} 
& \textbf{AutoComp} 
& \textbf{FlashInfer-Bench}
& \textbf{Agentic CUDA} 
& \textbf{Kernel Agent} 
& \textbf{Kernel Forge} \\
& \cite{astra2025} 
& \cite{wang2025geak} 
& \cite{cudaforge2025} 
& \cite{hong2025autocomp} 
& \cite{flashinferbench2026}
& \cite{lange2025robustkbench} 
& \cite{kernelagent2025} 
& \\
\hline

Uses LLM-based kernel generation and optimization
& \yes & \yes & \yes & \yes & \yes & \yes & \yes & \yes \\
\hline

Inserts generated kernels back into an execution path so the optimized workload can run end-to-end without manual kernel reintegration
& \no & \partialmark & \partialmark & \no & \yes & \partialmark & \partialmark & \yes \\
\hline

Uses tensors, shapes, arguments, and values captured from real model executions rather than randomly generated or standalone benchmark inputs
& \partialmark & \no & \no & \no & \partialmark & \partialmark & \no & \yes \\
\hline

Supports workloads beyond LLM-only serving or isolated benchmark kernels
& \no & \partialmark & \partialmark & \yes & \no & \partialmark & \partialmark & \yes \\
\hline

Provides a graphical interface to inspect kernels, track progress, compare candidates, and debug failures
& \no & \no & \no & \partialmark & \no & \no & \partialmark & \yes \\
\hline

\end{tabular}
  }
\end{table*}

Prior work closest to \tool{} falls into four broad groups: task-level kernel-generation benchmarks, DSL and workflow benchmarks, backend or production-integration systems, and autonomous optimization agents. First, task-level kernel and CUDA-code generation benchmarks such as KernelBench, KernelBench-X, robust-kbench, ComputeEval, CUDABench, and MultiKernelBench evaluate whether LLMs can produce correct and fast implementations for curated CUDA, Triton, or multi-platform tasks~\cite{ouyang2025kernelbench,wang2026kernelbenchx,lange2025robustkbench,nvidia2025computeeval,zhu2026cudabench,wen2025multikernelbench}. These suites are important for measuring model capability, correctness filtering, performance evaluation, and benchmark robustness. Kernel Forge is complementary: rather than starting from isolated benchmark prompts, it starts from traced model executions, groups concrete runtime variants, weights variants by observed runtime responsibility, and reports fallback behavior.

Second, DSL and workflow benchmarks such as TritonBench and TritonGym focus on Triton generation and agentic tool use~\cite{li2025tritonbench,guan2025tritongym}. These systems evaluate code-generation capability in a portable kernel language or standardized agent environment. Kernel Forge instead targets CUDA candidates for captured \torch{} variants and measures them against the actual \eager{} dispatch path observed in the workload.

Third, backend and production-oriented efforts such as BackendBench and FlashInfer-Bench evaluate whether generated kernels can be integrated into framework backends or deployed in inference-serving paths~\cite{backendbench2025,flashinferbench2026}. Kernel Forge shares the concern for integration, but emphasizes auditability: the evaluation distinguishes generated CUDA from wrapper or backend paths, records fallback decisions, and reports whether mixed replacement policies remain within the operator-region evidence boundary. This separation is important because an exported package can improve a captured operator-region outcome while still relying on fallback or wrapper-like behavior for some operators.

Fourth, autonomous kernel-generation and optimization systems such as Astra, CudaForge, AutoComp, GEAK, KernelAgent, and KernelFalcon study iterative search loops that use correctness tests, profiling, hardware feedback, specialized agent roles, or learned revision policies to produce optimized kernels or accelerator code~\cite{astra2025,cudaforge2025,hong2025autocomp,wang2025geak,kernelagent2025,kernelfalcon2025}. Kernel Forge also includes a revision-tree controller, but this paper does not claim search-policy superiority. Its emphasis is on measurement and packaging: trace weighting, per-variant provenance, correctness filters, fallback accounting, and conservative interpretation of guarded trace-weighted timing.

\tool{} also builds on compiler and runtime systems for deep learning. \torch{} provides optimized imperative execution and dispatches many CUDA operations through mature framework and vendor-backed paths such as ATen, cuBLAS, cuDNN, and other backend libraries~\cite{paszke2019pytorch}. TVM, AutoTVM, Ansor, and Triton generate efficient tensor programs through compiler IRs, schedules, templates, and measured or learned cost models~\cite{chen2018tvm,chen2018autotvm,zheng2020ansor,tillet2019triton}. FlashAttention and \torch{} SDPA illustrate why generated kernels must compete with strong backend implementations rather than naive references~\cite{dao2022flashattention,pytorch_sdpa_docs}.


\section{Conclusion}
\label{sec:conclusion}

We presented \textsc{Kernel Forge}, an open-source agentic AI harness for applying LLM-based CUDA kernel optimization to real PyTorch model executions. Rather than optimizing only isolated benchmark kernels, \textsc{Kernel Forge} captures operators from full model runs, groups them into workload-specific variants, generates and validates CUDA candidates, explores multiple optimization paths with MCTS, and exports guarded packages that fall back to PyTorch eager execution when replacement is not supported by the captured measurements. Across vision, diffusion, and LLM workloads, \textsc{Kernel Forge} finds operator-level improvements over PyTorch eager, reaching up to $2.83\times$ speedup on selected captured operators. At the same time, our results show why deployment-aware evaluation is important: the largest local speedups often occur on operators with small runtime contribution, while dominant operators are frequently backed by mature framework or vendor implementations. These findings suggest that agentic kernel optimization should be evaluated not only by whether it can generate fast standalone kernels, but also by how those kernels behave when applied inside real model executions with guarded replacement and workload-aware measurement.

\bibliographystyle{IEEEtran}
\bibliography{reference}

\end{document}